\documentclass[11pt,letterpaper]{article}
\usepackage{naaclhlt2016}
\usepackage{times}
\usepackage{latexsym}
\usepackage{graphicx}
\usepackage[english]{babel}
\usepackage{csquotes}
\usepackage{wrapfig}
\naaclfinalcopy 
\def\naaclpaperid{***} 

\title{Testing the limits of unsupervised learning for semantic similarity}

\author{Richa Sharma\\
	    Paralleldots, Inc.\\
	    {\tt richa@paralleldots.com}
	  \And
	Muktabh Srivastava\\
  	Paralleldots, Inc.\\
  {\tt muktabh@paralleldots.com}}

\date{}

\begin{document}

\maketitle

\begin{abstract}
 Semantic Similarity between two sentences can be defined as a way to determine how related or unrelated two sentences are. The task of Semantic Similarity in terms of distributed representations can be thought to be generating sentence embeddings (dense vectors) which take both context and meaning of sentence in account.
Such embeddings can be produced by multiple methods, in this paper we try to evaluate LSTM auto encoders for generating these embeddings. Unsupervised algorithms (auto encoders to be specific) just try to recreate their inputs, but they can be forced to learn order (and some inherent meaning to some extent) by creating proper bottlenecks. We try to evaluate how properly can algorithms trained just on plain English Sentences learn to figure out Semantic Similarity, without giving them any sense of what meaning of a sentence is.

\end{abstract}

\section{Introduction}

While good progress has been accomplished in generating the linguistic representations of individual words as dense vectors using plain text data (word embeddings), satisfactory representations of phrases and sentences in unsupervised learning is still a big challenge. In the study, we have focused on LSTM Auto encoders which on being properly tuned with the correct set of hyper parameters can have significant improvement in the working of the neural architecture to retrieve the useful semantic information. The LSTM auto encoders have the strong potential to exploit the context of the sentence if proper bottlenecks are used. A simple LSTM Auto encoder forms the baseline of our experiments, which can improve or degrade the results depending on the choice of parameters and bottleneck. We illustrate what crucial factors can affect the results of  simple auto encoder to generate semantic information in a positive way. While it has been shown that supervised models trained on a closely related task of NLI (Natural Language Inference which can be called a more detailed variant of semantic similarity) outperforms all unsupervised models, we wish to see how good we can get using unsupervised models (specifically LSTM auto encoders) only.

\section{Related Work}
Many architectures had been quite successful in generating the sentence embeddings using the unsupervised learning of word embeddings \cite{Pennington2014GloVe:Representation},\cite{Mikolov2013DistributedCompositionality}, \cite{Lin2017AEMBEDDING} as  well  as  supervised learning  of  sentence  embeddings cited in \cite{Conneau2017SupervisedData}. Following the existing work on learning word embeddings, most current approaches consider learning sentence encoders in an unsupervised manner like SkipThought \cite{Kiros2015Skip-ThoughtVectors} or \cite{Pagliardini2017UnsupervisedFeatures} and these get good accuracies on tasks like clustering or making t-SNE of the document. However, they have not been much effective in the task of calculating similarity of sentences with similar words as they are order agnostic. Our architecture uses the Long Short-Term Memory \cite{Hochreiter1997LONGMEMORY} auto encoder as LSTM is undoubtedly the successful feature extractor among all networks. We demonstrate that with the choice of right parameters, the model can capture the semantic context of the sentence.

\section{Models}
Our work proposes and compares the three variations of Long Short-Term Memory(LSTM) Auto encoder for the semantic similarity task. The simple unsupervised model in Figure 1 composes the sentence embeddings using word vectors. For a sequence of N words (w$_1$,.....,w$_{N}$), the base model computes a set of N hidden representations h$_1$,....,h$_{N}$ with h$_{t}$=LSTM(w$_1$,.....,w$_{N}$). The sentence is represented by the last hidden state h$_{N}$. This encoded sentence representation is referred to as  the Context Vector for easy recollection in this paper . The Context Vector initializes the hidden state of the decoder LSTM with starting input token. The decoder output is passed to the dense layer to predict the sequence of source words one by one from the context vector.  
\begin{figure}[h]
\begin{center}
\includegraphics[width=8cm]{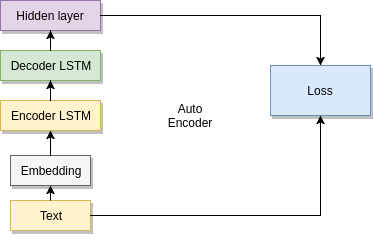}
\caption{\label{fig:fig1}Simple autoencoder.}
\end{center}
\end{figure}

\subsection{Basic Auto Encoder}
We have used previously trained word embeddings of Google News \footnote{https://github.com/3Top/word2vec-api} as the input to the unidirectional LSTM encoder with number of hidden units double than the input size to extract the sufficient information usefully. The encoded information is decoded by another unidirectional LSTM and words are predicted by the dense layer. This approach ( represented in the Figure 2 )helps to capture the semantic relatedness by narrowing the distance between the similar sentences but broadening the distance between the dissimilar sentences.
\begin{figure}[h]
\begin{center}
\includegraphics[width=8cm]{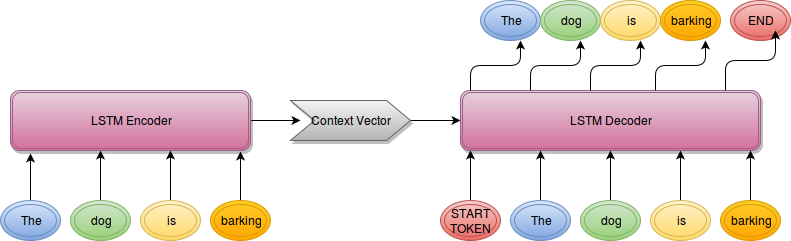}
\caption{\label{fig:fig2}Auto encoder with start token and end token.}
\end{center}
\end{figure}

\subsection{Cross Entropy and KL Divergence (KLD) in reconstruction Loss}

While replacing Cross Entropy loss with KL Divergence doesn’t train well, on combining Cross Entropy and KL Divergence as shown in Figure 3, we see something surprising. The loss function here is the aggregated result of the reconstruction loss and the Kullback-Leibler divergence (KLD) between the source distribution and the modeled distribution. This information divergence metric measures how much information is lost in fitted model relative to reference model. The objective of the model is to minimize weighted sum of the reconstruction and KLD loss. This KLD loss function results in significant improvement in determination of the similarity between the input and output of the auto encoder by narrowing down the distance between contextually similar words in a pair of sentences while modeling the source distribution to the target distribution.
\begin{figure}[h]
\begin{center}
\includegraphics[width=8cm]{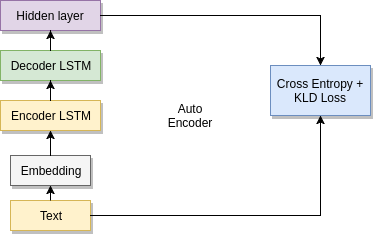}
\caption{\label{fig:fig2}Combined cross entropy and KLD loss autoencoder.}
\end{center}
\end{figure}
\subsection{Neural Variational Inference }

Since, the performance of auto encoder has been boosted with the use of KLD in the above model, we have experimented the slight variation in the auto encoder by evaluating the KLD between the context vector and  its Normal Distribution (with mean=0 and STD=1). The KLD between the context vector and its Normal distribution measures the amount of information gain from the source context vector and hence aim of the model is to minimize the combined result of the reconstruction loss, KLD on the context vector and KLD loss between the source distribution and the modeled distribution. Figure 4 represents the architecture of this proposed model.

\begin{figure}[h]
\begin{center}
\includegraphics[width=8cm]{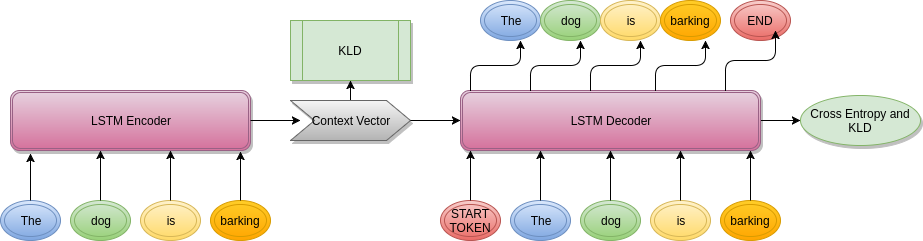}
\caption{\label{fig:fig2} Neural Variational Inference autoencoder.}
\end{center}
\end{figure}

\section{Results }
We  trained our models using SGD optimizer with learning rate of 0.1 and performed unsupervised evaluation of the learnt sentence embeddings in all the three models using the sentence  cosine  similarity, on the  STS 2014 \cite{Cer2017SemEval-2017Evaluation} and SICK 2014 \cite{Marelli2014SemEval-2014Entailment} datasets. These similarity scores are compared to the human-labeled similarity score using Pearson(Pearson, 1895) correlation scores. The SICK dataset consists of about 10,000 sentence pairs along with relatedness scores of the pairs between 0 and 5 and the STS 2014 dataset contains 3,770 pair of sentences. We have obtained pearson correlation scores for our models along with that of supervised model \cite{Conneau2017SupervisedData} as displayed in Table 1.  

\begin{table}
\small
\centering
\begin{tabular}{|p{3cm}|p{2cm}| p{2cm}|}
\hline \bf Type of Model & \bf Correlation (STS) & \bf Correlation (SICK)  \\ \hline
Basic Autoencoder & 0.21 & 0.23  \\
Variational Autoencoder & 0.14 & 0.15  \\
Cross Entropy and KLD in reconstruction Loss & 0.34 & 0.37\\
Supervised \cite{Conneau2017SupervisedDatab} & 0.68 & -\\

\hline
\end{tabular}
\caption{\label{font-table} Comparison of the performance of different models.}
\end{table}

\section{Conclusion}

This paper does not offer a practical methodological approach beyond establishing reliable baselines but studies the effects of hyperparameter optimization strategies to learn the generic sentence embeddings in unsupervised way. By exploring different architectures on large datasets, we demonstrated that LSTM with KLD could enhance the quality of the learnt sentence embeddings.
\bibliography{Mendeley}
\bibliographystyle{naaclhlt2016}
\end{document}